# Deep Reinforcement Learning for *de-novo* Drug Design

*Mariya Popova[1,2,3], Olexandr Isayev[1*], Alexander Tropsha[1*]*

[1]Laboratory for Molecular Modeling, Division of Chemical Biology and Medicinal Chemistry, UNC Eshelman School of Pharmacy, University of North Carolina, Chapel Hill, NC, 27599, USA.

[2]Moscow Institute of Physics and Technology, Dolgoprudny, Moscow region, 141700, Russia.

[3]Skolkovo Institute of Science and Technology, Moscow, 143026, Russia.

*Correspondence: A.T (alex_tropsha@unc.edu), O.I. (olexandr@olexandrisayev.com).

## Abstract

We propose a novel computational strategy for *de-novo* design of molecules with desired properties termed ReLeaSE (Reinforcement Learning for Structural Evolution). Based on deep and reinforcement learning approaches, ReLeaSE integrates two deep neural networks – generative and predictive – that are trained separately but employed jointly to generate novel targeted chemical libraries. ReLeaSE employs simple representation of molecules by their SMILES strings only. Generative models are trained with stack-augmented memory network to produce chemically feasible SMILES strings, and predictive models are derived to forecast the desired properties of the *de novo* generated compounds. In the first phase of the method, generative and predictive models are trained separately with a supervised learning algorithm. In the second phase, both models are trained jointly with the reinforcement learning approach to bias the generation of new chemical structures towards those with the desired physical and/or biological properties. In the proof-of-concept study, we have employed the ReLeaSE method to design chemical libraries with a bias toward structural complexity or biased toward compounds with either maximal, minimal, or specific range of physical properties such as melting point or hydrophobicity, as well as to develop novel putative inhibitors of JAK2. The approach proposed herein can find a general use for generating targeted chemical libraries of novel compounds optimized for either a single desired property or multiple properties.

## Introduction

The combination of big data and artificial intelligence was referred to by the World Economic Forum as the fourth industrial revolution, that can radically transform the practice of scientific discovery (*1*). Artificial intelligence (AI) is revolutionizing medicine (*2*) including radiology, pathology and other medical specialties (*3*). Deep Learning (DL) technologies are beginning to find applications in drug discovery (*4*, *5*) including areas of molecular docking (*6*) , transcriptomics (*7*) , reaction mechanism elucidation (*8*) , and molecular energy prediction (*9*, *10*).

The crucial step in many new drug discovery projects is the formulation of a well-motivated hypothesis for new lead compound generation (*de novo* design) or compound selection from available or synthetically feasible chemical libraries based on the available SAR data. The design hypotheses are often biased towards preferred chemistry (*11*) or driven by model interpretation (*12*). *Automated* approaches for designing compounds with the desired properties *de novo* have become an active field of research in the last 15 years (*13–15*). The diversity of synthetically feasible chemicals that can be considered as potential drug-like molecules was estimated to be between $10^{30}$ and $10^{60}$ (*16*). Great advances in computational algorithms (*17*, *18*), hardware, and high-throughput screening (HTS) technologies (*19*) notwithstanding, the size of this virtual library prohibits its exhaustive sampling and testing by systematic construction and evaluation of each individual compound. Local optimization approaches have been proposed but they do not ensure the optimal solution, as the design process converges on a local or 'practical' optimum by stochastic sampling, or restricts the search to a defined section of chemical space which can be screened exhaustively (*13*, *20*, *21*).

Notably, a method for exploring chemical space based on continuous encodings of molecules was proposed recently (*22*). It allows efficient, directed gradient-based search through chemical space but does not include biasing libraries toward special physical or biological properties. Another very recent approach for generating focused molecular libraries with the desired bioactivity using Recurrent Neural Networks was proposed as well (*23*). However, properties of produced molecules could not be controlled well. Adversarial autoencoder was proposed recently (*24*) as a tool for generating new molecules with the desired properties; however, compounds of interest are selected by the means of virtual screening of large libraries, not by designing novel molecules. Specifically, in this method, points from the latent space of chemical descriptors are projected to the nearest known molecule in the screening database.

Herein, we propose a novel method for generating chemical compounds with desired physical, chemical and/or bioactivity properties *de novo* that is based on deep reinforcement learning (RL). Reinforcement learning is a subfield of artificial intelligence, which is used to solve dynamic decision problems. It involves the analysis of possible actions and estimation of the statistical relationship between the actions and their possible outcomes, followed by the determination of a treatment regime that attempts to find the most desirable outcome. The integration of reinforcement learning and neural networks dates back to 1990s (*25*). However, with the recent advancement of deep learning (DL), benefiting from Big Data, new powerful algorithmic approaches are emerging. There is a current renaissance of RL (*26*), especially when it is combined with deep neural networks, i.e., deep reinforcement learning. Most recently, RL was employed to achieve superhuman performance in the game Go (*27*), considered practically intractable due to the theoretical complexity of over $10^{140}$ possible solutions (*28*). One may see an analogy with the complexity of chemical space exploration with an algorithm that avoids brute-force computing to examine every possible solution. Below we describe the application of deep RL to the problem of designing chemical libraries with the desired properties and show that our approach termed ReLeaSE affords plausible solution of this problem.

The proposed ReLeaSE approach alleviates the deficiency of a small group of methodologically similar approaches discussed above. The most distinct innovative aspects of the approach proposed herein include the simple representation of molecules by their SMILES strings only for both generative and predictive phases of the method and integration of these phases into a single workflow including RL module. We demonstrate that ReLeaSE enables the design of chemical libraries with the desired physico-chemical and biological properties. Below we discuss both the algorithm and its proof-of-concept applications to designing targeted chemical libraries.

## Results

We have devised a novel RL-based method termed ReLeaSE (Reinforcement Learning for Structural Evolution) for generating new chemical compounds with desired physical, chemical or bioactivity properties. The general workflow for the ReLeaSE method (Figure 1) includes two deep neural networks (generative *G* and predictive

*P*). The process of training consists of two stages. During the first stage, both models are trained separately with supervised learning algorithms, and during the second stage, the models are trained jointly with a reinforcement learning approach that optimizes target properties. In this system, the generative model is used to produce novel chemically feasible molecules, i.e., it plays a role of an agent whereas the predictive model (that predicts properties of novel compounds) plays the role of a critic, which estimates the agent's behavior by assigning a numerical reward (or penalty) to every generated molecule. The reward is a function of the numerical property generated by the predictive model, and the generative model is trained to maximize the expected reward.

**Reinforcement learning formulation as applied to chemical library design.** Both generative (*G)* and predictive (*P)* models are combined into a single RL system. The set of actions *A* is defined as an alphabet, i.e., the entire collection of letters and symbols is used to define canonical SMILES strings that are most commonly employed to encode chemical structures. For example, an aspirin molecule is encoded as [CC(=O)OC1=CC=CC=C1C(=O)O]. The set of states *S* is defined as all possible strings in the alphabet with lengths from zero to some value *T*. The state $s_0$ with length 0 is unique and considered the initial state. The state $s_T$ of length *T* is called the terminal state, as it causes training to end. The subset of terminal states $S^* = \{s_T \in S\}$ of *S,* which contains all states $s_T$ with length *T* is called the terminal states set. Reward $r(s_T)$ is calculated at the end of the training cycle when the terminal state is reached. Intermediate rewards $r(s_t),\ t < T$ are equal to zero. In these terms, the generative network *G* can be treated as a policy approximation model. At each time step *t, 0 < t < T*, *G* takes the previous state $s_{t-1}$ as an input and estimates probability distribution $p(a_t \mid s_{t-1})$ of the next action. Afterwards, the next action $a_t$ is sampled from this estimated probability. Reward $r(s_T)$ is a function of the predicted property of $s_T$ using the predictive model *P*:

$$r(s_T) = f\big(P(s_T)\big) \tag{1}$$

where *f* is chosen depending on the task. Some examples of the functions *f* are provided in the computational experiment section. Given these notations and assumptions, the problem of generating chemical compounds with desired properties can be formulated as a task of finding a vector of parameters *Θ* of policy network *G* which maximizes the expected reward:

$$J(\Theta) = \mathbb{E}[r(s_T) \mid s_0, \Theta] = \sum_{s_T \in S^*} p_\Theta(s_T) r(s_T) \ \rightarrow max. \tag{2}$$

This sum iterates over the set $S^*$ of terminal states. In our case, this set is exponential and the sum cannot be computed exactly. According to the law of large numbers, we can approximate this sum as a mathematical expectation by sampling terminal sequences from the model distribution:

$$J(\Theta) = \mathbb{E}[r(s_T) \mid s_0, \Theta] = \mathbb{E}_{a_1 \sim p_\Theta(a_1 \mid s_0)} \mathbb{E}_{a_2 \sim p_\Theta(a_2 \mid s_1)} \dots \mathbb{E}_{a_T \sim p_\Theta(a_T \mid s_{T-1})} r(s_T). \tag{3}$$

To estimate *J(Θ),* we sequentially sample $a_t$ from the model *G* for *t* from 0 to *T*. The unbiased estimation for *J(Θ)* is the sum of all rewards in every time step, which, in our case, equals the reward for the terminal state as we assume that intermediate rewards are equal to 0. This quantity needs to be maximized; therefore, we need to compute its gradient. This can be done, for example, with the REINFORCE algorithm (*29*) that uses the approximation of mathematical expectation as a sum, which we provided in equation 3, and the following form:

$$\partial_\Theta f(\Theta) = f(\Theta)\frac{\partial_\Theta f(\Theta)}{\partial \Theta} = f(\Theta)\partial_\Theta \log f(\Theta). \tag{4}$$

Therefore, the gradient of *J(Θ)* can be written down as:

$$\partial_\Theta J(\Theta) \ = \sum_{s_T \in S^*} [\partial_\Theta p_\Theta(s_T)] r(s_T) = \sum_{s_T \in S^*} p_\Theta(s_T)[\partial_\Theta \log p_\Theta(s_T)] r(s_T) \tag{5}$$

$$= \sum_{s_T \in S^*} p_\Theta(s_T)\left[\sum_{t=1}^{T} \partial_\Theta \log p_\Theta(a_t|s_{t-1})\right] r(s_T)$$

$$= \mathbb{E}_{a_1 \sim p_\Theta(a_1|s_0)} \mathbb{E}_{a_2 \sim p_\Theta(a_2|s_1)} \dots \mathbb{E}_{a_T \sim p_\Theta(a_T|s_{T-1})} \left[\sum_{t=1}^{T} \partial_\Theta \log p_\Theta(a_t|s_{t-1})\right] r(s_T),$$

which gives an algorithm $\partial_\Theta J(\Theta)$ estimation.

**Neural networks architectures.** Model *G* (Figure 1A) is a generative recurrent neural network, which outputs molecules in SMILES notation. We use a special type stack-augmented recurrent neural network (Stack-RNN) (*30*) that has found success in inferring algorithmic patterns. In our implementation, we consider legitimate (i.e., corresponding to chemically feasible molecules) SMILES strings as sentences composed of characters used in SMILES notation. The objective of Stack-RNN then is to learn hidden rules of forming sequences of letters that correspond to legitimate SMILES strings.

To generate a valid SMILES string, in addition to correct valence for all atoms, one must count, ring opening and closure, as well as bracket sequences with several bracket types. Regular recurrent neural networks like LSTM (*31*) and GRU (*32*) are unable to solve the sequence prediction problems due to their inability to count. One of the classical examples of sequences that cannot be properly modeled by regular recurrent networks are words from the Dyck language, where all open square brackets are matched with the respective closed ones (*33*). Formal language theory states that context-free languages, such as Dyck language, can not be generated by model without stack memory (*34*). As valid SMILES string should at least be a sequence of all properly matched parentheses with multiple types of brackets, recurrent neural networks with an additional memory stack is a theoretically justified choice for modelling SMILES. Another weakness of regular recurrent neural networks is their inability to capture long term dependencies, which leads to difficulties in generalizing to longer sequences (*35*). All of these features are required to learn the language of the SMILES notation. In a valid SMILES molecule, in addition to correct valence for all atoms, one must count, ring opening and closure, as well as bracket sequences with several bracket types. Therefore, only memory-augmented neural networks like Stack-RNN or Neural Turing Machines are the appropriate choice for modeling such sequence dependencies.

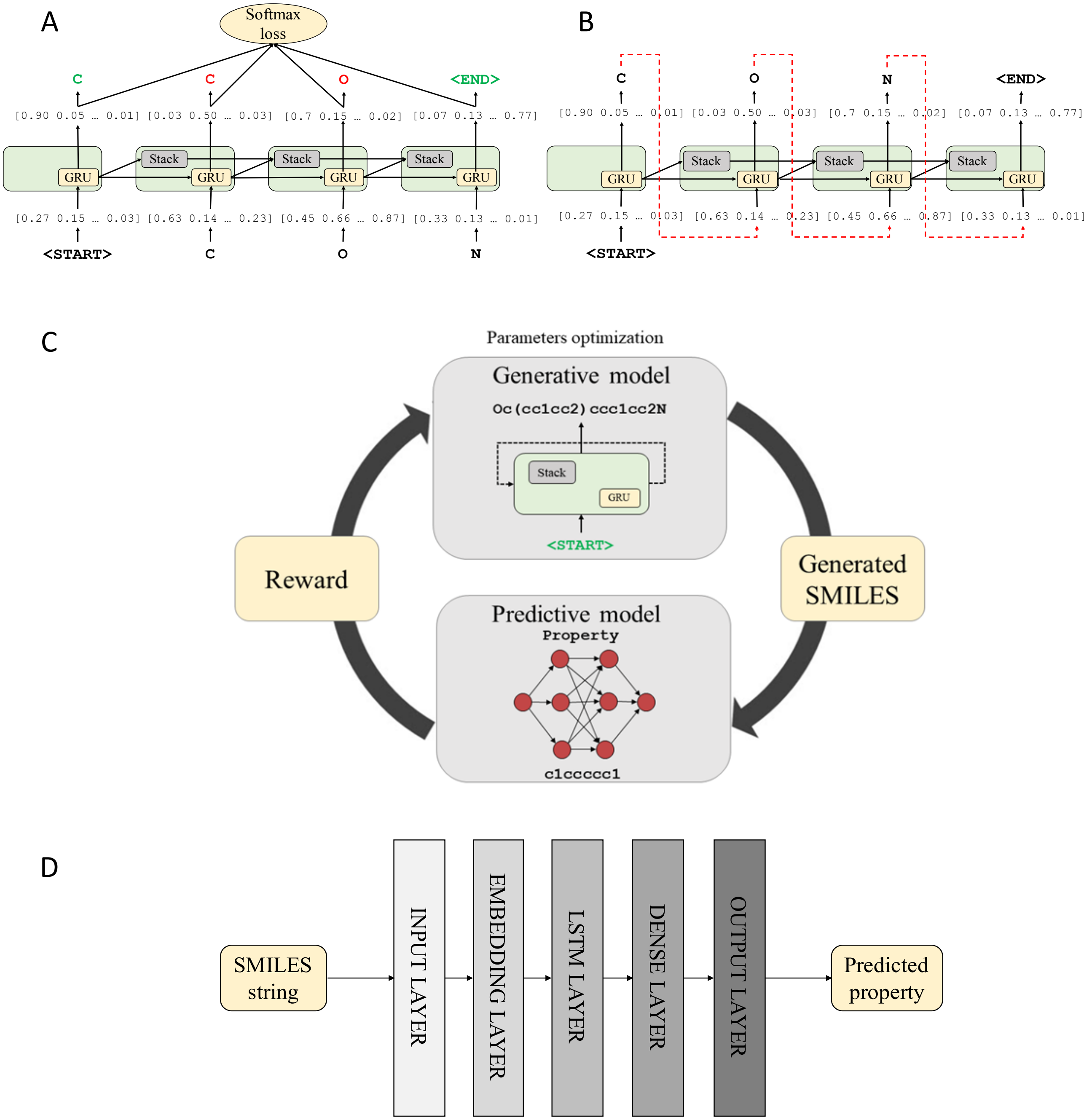

**Figure 1. The workflow implementingdDeep RL algorithm for generating new SMILES strings of compounds with the desired properties. A**. Training step of the generative stack-augmented RNN; **B**. Generator step of the generative stack-augmented RNN. During training, the input token is a character in the currently processed SMILES string from the training set. The model outputs the probability vector $p_\Theta(a_t|s_{t-1})$ of the next character given a prefix. Vector of parameters $\Theta$ is optimized by cross-entropy loss function minimization. In the generator regime, the input token is a previously generated character. Next, character $a_t$ is sampled randomly from the distribution $p_\Theta(a_t|s_{t-1})$. **C.** General pipeline of reinforcement learning system for novel compounds generation. **D.** Scheme of predictive model. This model takes a SMILES string as an input and provides one real number, which is an estimated property value, as an output. Parameters of the model are trained by $l_2$ squared loss function minimization.

The Stack-RNN defines a new neuron or cell structure on top of the standard GRU cell (see Figure 1A). It has two additional multiplicative gates referred to as the memory stack, which allow the Stack-RNN to learn meaningful long-range interdependencies. Stack is a differentiable structure onto and from which continuous vectors are inserted and removed. In stack terminology, the insertion operation is called `PUSH` operation and the removal operation is called `POP` operation. These traditionally discrete operations are continuous here, since `PUSH` and `POP` operations are permitted to be real values in the interval (0, 1). Intuitively, we can interpret these values as the degree of certainty with which some controller wishes to `PUSH` a vector $v$ onto the stack, or `POP` the top of the stack. Such an architecture resembles a *pushdown automata*, which is a classic framework from the theory of formal languages capable of dealing with more complicated languages. Applying this concept to neural networks provides the possibility to build a trainable model of the language of SMILES with correct syntaxes, proper balance of ring-opening and closures and correct valences for all elements.

The second model $P$ is a predictive model (see Figure 1D) for estimating physical, chemical or biological properties of molecules. This property prediction model is a deep neural network, which consists of an embedding layer, LSTM layer and two dense layers. This network is designed to calculate user-specified property (activity) of the molecule taking SMILES string as an input data vector. In a practical sense, this learning step is analogous to traditional Quantitative Structure-Activity Relationships (QSAR) models. However, unlike conventional QSAR, no numerical descriptors are needed, as the model distinctly learns directly from the SMILES notation as to how to relate the comparison between SMILES strings to that between target properties.

**Generation of chemicals with novel structures.** The generative network was trained with ~1.5M structures from ChEMBL21 database (*36*) (please see Methods for technical details); the objective of training was to learn rules of organic chemistry that define SMILES string corresponding to realistic chemical structures. To demonstrate the versatility of the baseline (unbiased) Stack-RNN, we generated over one million (1M) of compounds. All structures are available for download from Supplementary Information. Random examples of the generated compounds are illustrated in Figure 2.

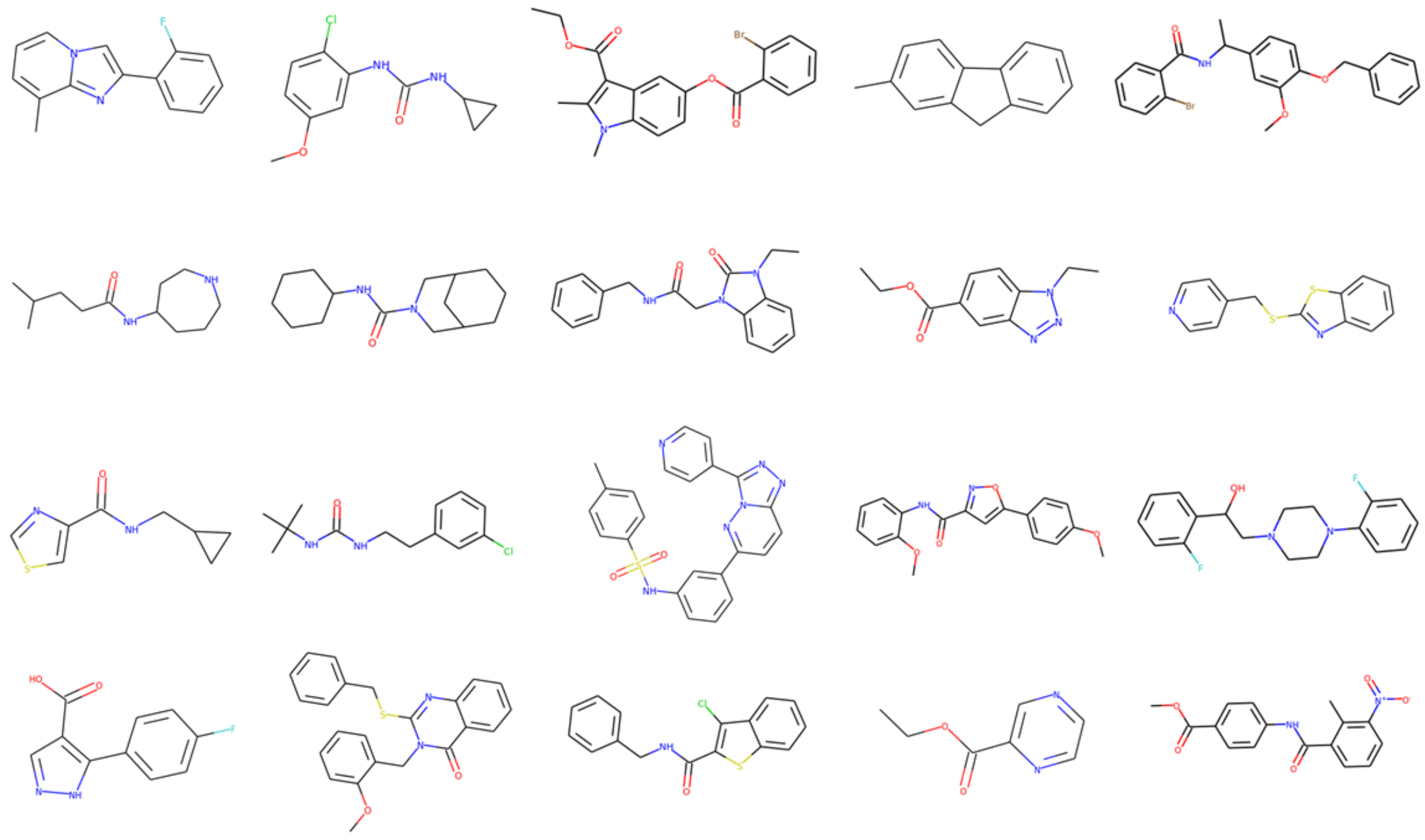

**Figure 2. A sample of molecules produced by the generative model**.

A known deficiency of approaches for *de novo* molecular design is frequent generation of chemically infeasible molecules.(*22*, *37*) To address this possible issue of concern, we have established that 95% of all generated structures, were valid, chemically sensible, molecules. The validity check was performed by the structure checker from ChemAxon (*38*). We compared the 1M de novo generated molecules with those used to train the generative model from the ChEMBL database and found that the model produced less than 0.1% of structures from the training dataset. Additional comparison with the ZINC15 database (*39*) of 320M synthetically accessible drug-like molecules showed that about 3% (~32,000 molecules) of *de novo* generated structures could be found in ZINC. All ZINC IDs for the matching molecules are available in the Supplementary Information.

To assess the importance of using stack memory augmented network as described in Methods, we compared several characteristics of chemical libraries generated by models developed either with or without stack memory. For this purpose, we trained another generative recurrent neural network with the same architecture but without using stack memory. Libraries were compared by the percentage of valid generated SMILES, internal diversity, and similarity of the generated molecules to those in the training dataset (ChEMBL). The model without stack memory showed that only 86% of molecules in the respective library were valid (as evaluated by ChemAxon, cf. Methods) compared to 95% of valid molecules in the library generated with stack memory network. As expected (cf. the justification for using stack memory augmented network in Methods), in the former library, syntactic errors such as open brackets, unclosed cycles and incorrect bond types in SMILES strings were more frequent. Based on the analysis of respective sets of 10000 molecules generated by each method (See Figure 3A), the library obtained without stack memory showed a decrease of the internal diversity by 0.2 units of the Tanimoto coefficient and yet, a four-fold increase in the number of duplicates, from just about 1% to 5%. In addition, Figure 3B shows that the number of molecules similar to the training dataset (Ts > 0.85) for library obtained without stack memory (28% of all molecules) is twice that obtained with stack memory (14%). These results clearly highlight the advantages of using neural network with memory for generating the highest number of realistic and predominantly novel molecules, which is one of the chief objectives of *de novo* chemical design.

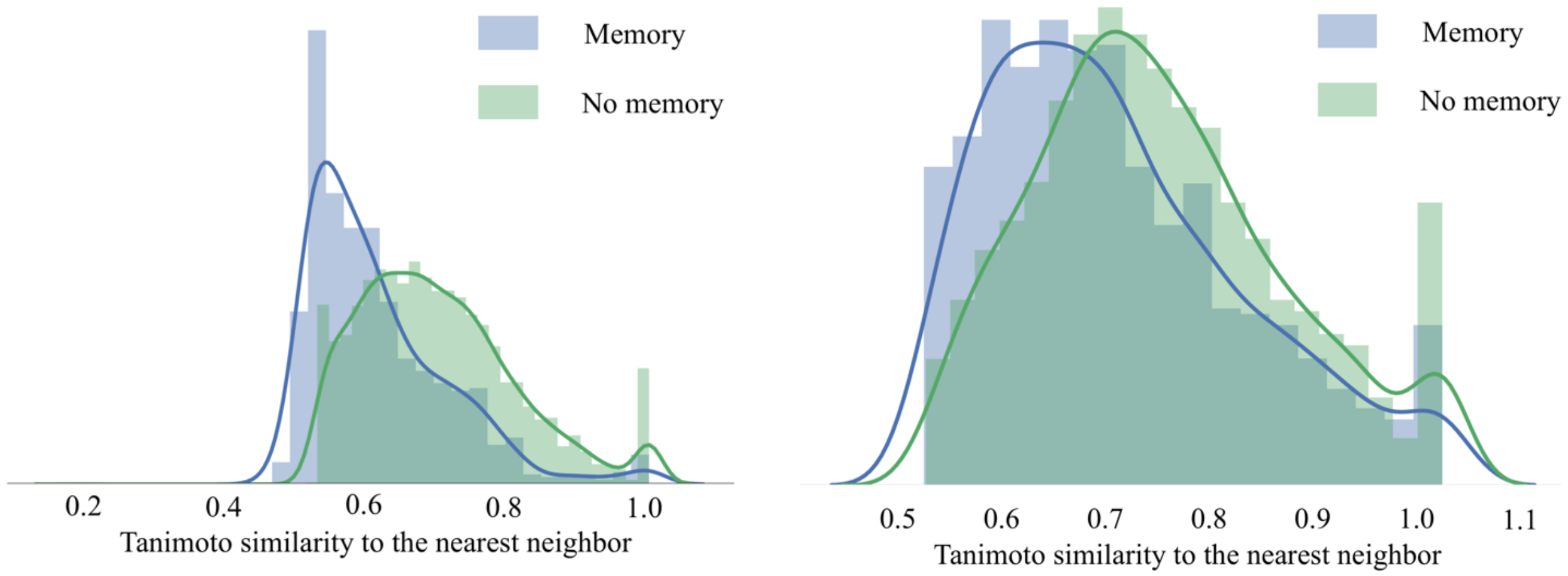

**Figure 3. Performance of the generative model G, with and without stack-augmented memory**. (**A**) Internal diversity of generated libraries; (**B**) Similarity of the generated libraries to the training dataset from the ChEMBL database.

In order to further characterize the structural novelty of the *de novo* generated molecules, we compared the content of the Murcko scaffolds (*40*) between the ChEMBL training set and the virtual library generated by our system. Murcko scaffolds provide a hierarchical molecular organization scheme by dividing small molecules into R-groups, linkers, and frameworks, or scaffolds. They define the ring systems of a molecule by removing side chain atoms. We found that less than 10% of scaffolds in our library were present in ChEMBL. Overall, this analysis suggests that generative Stack-RNN model did not simply memorize the training SMILES sequences but was indeed capable of generating extremely diverse yet realistic molecules as defined by the structure checker from ChemAxon.

In addition to passing the structure checker, an important requirement for *de novo* generated molecules is their synthetic feasibility. To this end, we employed the synthetic accessibility score (SAS) method (*41*), which relies on the knowledge extracted from known synthetic reactions and adds penalty for high molecular complexity. For ease of interpretation, SAS is scaled to be between 1 and 10. Molecules with the high SAS value, typically above 6 are considered difficult to synthesize, whereas, molecules with the low SAS values are easily synthetically accessible. The distribution of SAS values calculated for 1M molecules generated by the ReLeaSE is shown in Supplementary Figure S1. To illustrate the robustness of the *de novo* generated chemical library, we compared its SAS distribution with that of the SAS values both for the full ChEMBL library (~1.5M molecules) and for 1M random sample of molecules in ZINC. Similar to typical commercial vendor libraries, distribution of SAS for ReLeaSE is skewed towards more easily synthesizable molecules. Median SAS values were 2.9 for ChEMBL and 3.1 for both ZINC and ReLeaSE. Over 99.5% of de novo generated molecules had SAS values below 6. Therefore, despite their high novelty, vast majority of generated compounds can be considered as synthetically accessible.

**Property prediction.** Over more than 50 years of active development of the field, well-defined QSAR protocols and procedures have been established (*42*), including best practices for model validation as reported in several highly cited papers by our group (*42*, *43*). Any QSAR method can be generally defined as an application of machine learning and/or statistical methods to the problem of finding empirical relationships of the form $y = f(X_1, X_2,\ldots,X_n)$, where $y$ is biological activity (or any property of interest) of molecules; $X_1, X_2,\ldots, X_n$ are calculated molecular descriptors of compounds; and, $f$ is some empirically established mathematical transformation that should be applied to descriptors to calculate the property values for all molecules. Model validation is a critical component of model development; our approach to model validation in this study is described in Methods.

Building machine learning (ML) models directly from SMILES strings, which is a unique feature of our approach, completely bypasses the most traditional step of descriptor generation in QSAR modeling. In addition to being relatively slow, descriptor generation is non-differentiable and it does not allow a straightforward inverse mapping from the descriptor space back to molecules albeit a few approaches for such mapping (i.e., inverse-QSAR) have been proposed (*44–46*). For instance, one of the studies described above (*22*) used mapping from the point in latent variable to real molecules represented by points most proximal to that point. In contrast, using neural networks directly on SMILES is fully differentiable, and it also enables direct mapping of properties to the SMILES sequence of characters (or strings). SMILES strings were used for QSAR model building previously (*47*, *48*); however, in most cases SMILES strings were used to derive string- and substring-based numerical descriptors (*49*). Note that, in our case, the ability to develop QSAR models using SMILES was critical for integrating property assessment (evaluative models) and *de novo* structure generation (generative models) into a single RL workflow as described below.

In terms of external prediction accuracy, SMILES based ML models also performed very well. For example, using five-fold cross validation we obtained the external model accuracy expressed as $R^2_{ext}$ of 0.91 and RMSE = 0.53 for predicting LogP (See Methods section). This compared favorably to a Random Forest model with DRAGON7 descriptors ($R^2_{ext}$ = 0.90 and RMSE = 0.57). For the melting temperature prediction, the observed RMSE of 35 $^oC$ was the same as that predicted with the state-of-the-art consensus model obtained by using an ensemble of multiple conventional descriptor-based ML models (*50*), which afforded RMSE of 35 $^oC$.

The following study was undertaken to evaluate the external predictive accuracy for novel compounds designed with the ReLeaSE method. We have identified over 100 compounds from our library in the ChEMBL database that were not present in the training set. Then we manually extracted their experimental LogP or Tmelt data from as recorded in PubChem database. Multiple measurements were averaged. Final subsets were comprised from about 20 molecules for each property. The comparison between predicted and experimental measurements yielded RMSE of 0.9 for LogP and ~42 degrees for Tmelt. This accuracy was slightly lower than that for the respective QSPR model obtained with cross-validation. We consider the reasonable success of this exercise in property prediction for an external dataset as an additional evidence that our approach yields molecules with both desired and accurately predicted properties

**Generation of property value biased libraries with the RL system.** To explore the utility of the RL algorithm in a drug design setting, we have conducted case studies to design libraries with three controlled target properties: a) physical properties considered important for drug-like molecules, b) specific biological activity, and c) chemical complexity. For physical properties, we selected melting temperature ($T_m$) and n-octanol/ water partition coefficient (LogP). For bioactivity prediction, we designed putative inhibitors of Janus protein kinase 2 (JAK2) with novel chemotypes. Finally, the number of benzene rings and the number of substituents (like –OH, $-NH_2$, $-CH_3$ –CN, etc.) was used as a structural reward to design novel chemically complex compounds. Figure 4 shows the distribution of predicted properties of interest in the training test molecules and in the libraries designed by our system. In all cases, we sampled 10,000 molecules by the baseline (no RL) generator and RL-optimized generative models, and then calculated their properties with a corresponding predictive model. Values of the substructural features were calculated directly from the 2D structure. Table 1 summarizes the analysis of generated molecules and the respective statistics.

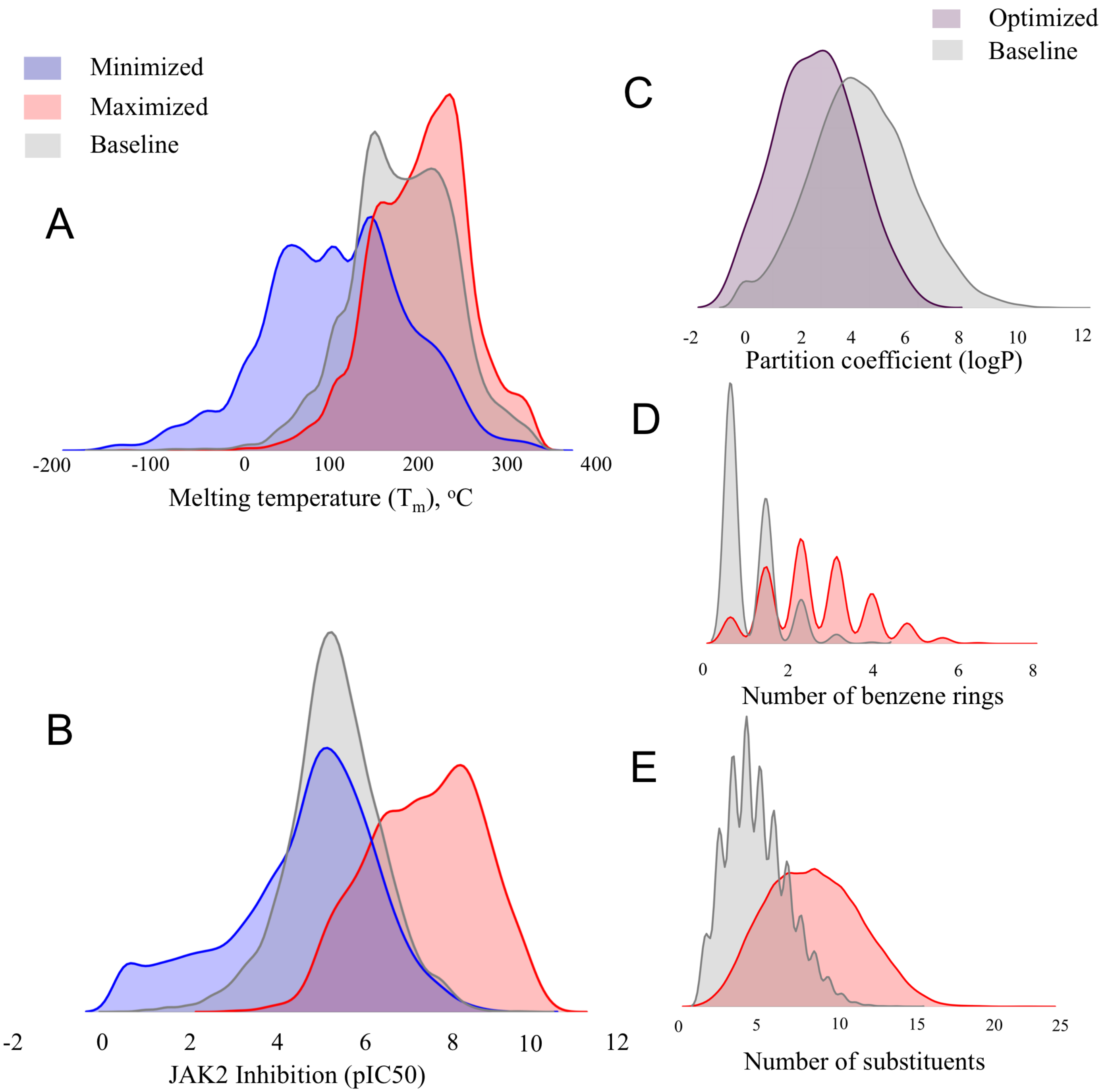

**Figure 4. Property distributions for RL optimized versus baseline generator model. (A)** Melting temperature; **(B)** JAK2 inhibition; **(C)** Partition coefficient; **(D)** Number of benzene rings; **(E)** Number of substituents.

*Melting temperature ($T_m$).* In this experiment, we set two goals, i.e., either to minimize or to maximize the target property. Upon minimization, the mean of the distribution in the *de novo* generated library was shifted by 44 $^{o}$C as compared to the training set distribution (Figure 4A). The library of virtually synthesized chemicals included simple hydrocarbons like butane, as well as poly-halogenated compounds like $CF_2Cl_2$ and $C_6H_4F_2$. The molecule with the lowest $T_m$=-184 °C in the produced dataset was $CF_4$. Clearly, this property minimization strategy was extremely effective, as it allowed for the discovery of molecules in the regions of the chemical space far beyond those of the training set of drug-like compounds. In the maximization regime, the mean of the melting temperature was increased by 20 °C to 200 °C. As expected, the generated library indeed included substantially more complex molecules with the abundance of sulphur-containing heterocycles, phosphates, and conjugated double bond moieties.

*Designing a chemical library biased toward a range of lipophilicity (LogP).* Compound hydrophobicity is an important consideration in drug design. One of the components of the famous Lipinski's rule of five is that orally bioavailable compounds should have their octanol-water partition coefficient LogP less than 5 (*51*). Thus, we endeavored to design a library that would contain compounds with LogP values within a favorable drug-like range. The reward function in this case was defined as a piecewise linear function of LogP with a constant region from 1.0 to 4.0 (see Supplementary Figure S2). In other words, we set the goal to generate molecules according to a typical Lipinski's constraint. As is shown in Figure 4C, we have succeeded in generating a library with 88% of the molecules falling within the drug-like region of LogP values.

*Inhibition of JAK2.* In the third experiment, which serves as an example of the most common application of computational modeling in drug discovery, we have employed our system to design molecules with the specific biological function, i.e., JAK2 activity modulation. Specifically, we designed libraries with the goal of minimizing or maximizing $pIC_{50}$ values for JAK2. While most of drug discovery studies are oriented toward finding molecules with heightened activity, bioactivity minimization is also pursued in drug discovery to mitigate off-target effects. Therefore, we were interested in exploring the ability of our system to bias the design of novel molecular structures toward any desired range of the target properties. JAK2 is non-receptor tyrosine kinase involved in various processes such as cell growth, development, differentiation or histone modifications. It mediates essential signaling events in both innate and adaptive immunity. In the cytoplasm it also plays an important role in signal transduction. Mutations in JAK2 have been implicated in multiple conditions like thrombocythemia, myelofibrosis or myeloproliferative disorders (*52*).

The reward functions in both cases (min and max) were defined as exponential functions of $pIC_{50}$ (see Supplementary Figure S2). The results of library optimization are shown in Figure 4B. With minimization, the mean of predicted $pIC_{50}$ distribution was shifted by about one $pIC_{50}$ unit and the distribution was heavily biased toward the lower ranges of bioactivity with 24% of molecules predicted to have practically no activity ($pIC_{50} \leq 4$). In the activity maximization exercise, properties of generated molecules were more tightly distributed across the predicted activity range. In each case, our system virtually synthesized both known and novel compounds, with the majority of *de novo* designed molecules being novel compounds. The generation of known compounds (i.e. not included in the training set) can be regarded as model validation. Indeed, the system retrospectively discovered 793 commercially available compounds deposited in the ZINC database, which constituted about 5% of the total generated library. Importantly, as many as 15 of them (exemplified by ZINC263823677 - http://zinc15.docking.org/substances/ZINC000263823677/ and ZINC271402431 - http://zinc15.docking.org/substances/ZINC000271402431/) were actually annotated as possible tyrosine kinase inhibitors.

*Substructure bias.* Finally, we also performed two simple experiments mimicking the strategy of biased chemical library design where the designed library is enriched with certain user-defined substructures. We defined the reward function as the exponent of (i) the number of benzene rings (-Ph) and (ii) total number of small group substituents. Among all case studies described, structure bias was found to be the easiest to optimize. The results of the library optimization study are shown in Figures 4 D, E. Furthermore, Figure 5 illustrates the evolution of generated structures as the structural reward increases. Indeed, we see that the model progresses toward generating increasingly more complex, yet realistic molecules with greater numbers of rings and/or substituents.

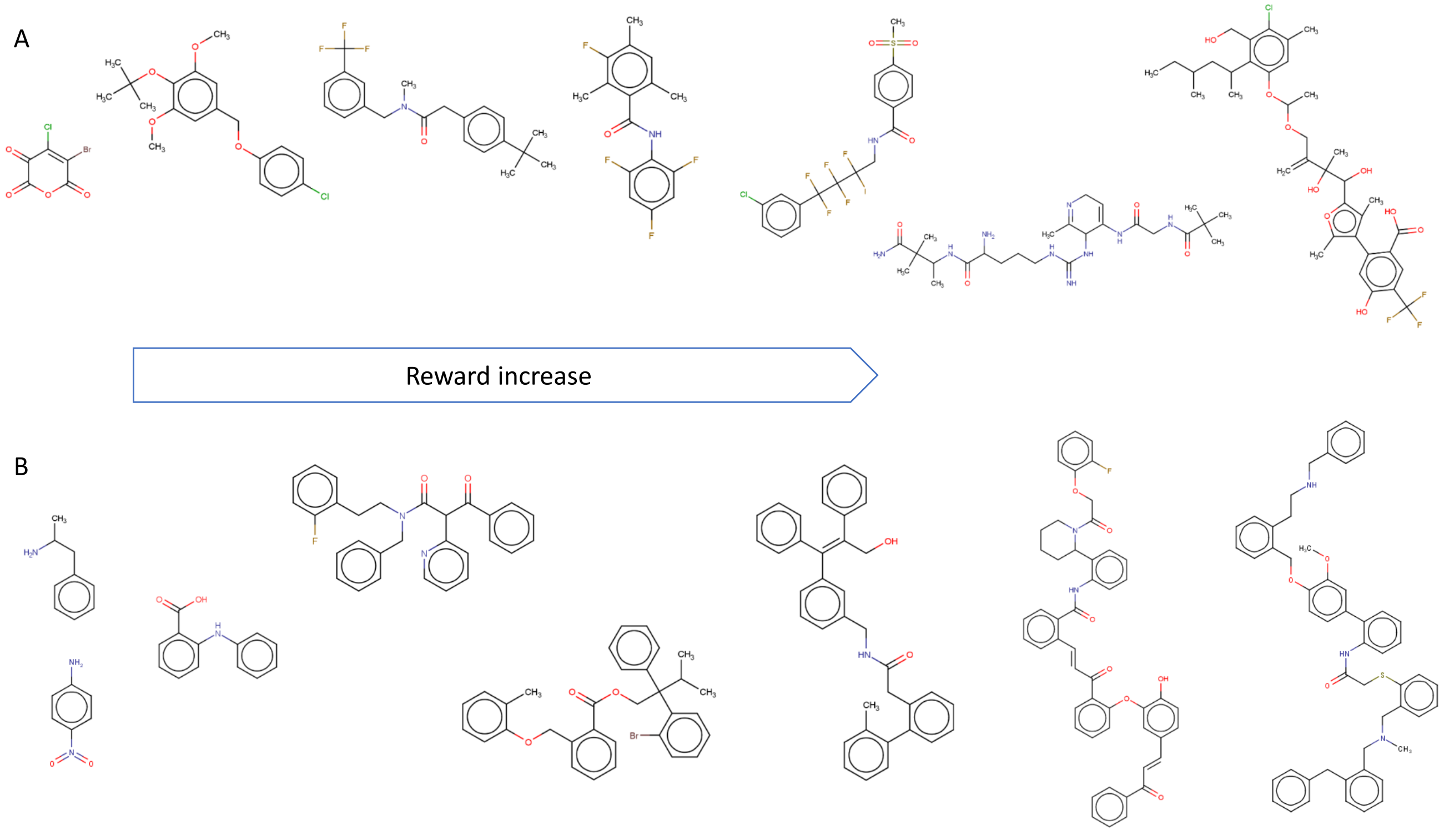

**Figure 5. Evolution of generated structures as chemical substructure reward increases**. (**A)** Reward proportional to the total number of small group substituents (**B**) Reward proportional to the number of benzene rings.

We expect that designing structurally biased libraries may be a highly desirable application of the ReLeaSE approach as researchers often wish to generate libraries enriched for certain privileged scaffold(-s) and lead compounds optimization (*53*). Conversely, the system also allows to avoid particular chemical groups or substructures (like bromine or carboxyl group) that may lead to undesired compound properties such as toxicity. Finally, one could implement certain substructure, or pharmacophore similarity, reward to explore additional chemical space.

Table 1 shows a decrease in the proportion of the valid molecules after the optimization. We may explain this phenomenon by the weaknesses of predictive models *P* (See Figure 1C) and the integration of predictive and generative models into a single design system. We presume that the generative model *G* tends to find some local optima of the reward function that correspond to invalid molecules, but predictive model *P* assigns high rewards to these molecules. This explanation is also supported by the results of structure bias optimization experiments, as we did not use any predictive models in these experiments and the decrease in the proportion of valid molecules was insignificant. We also noticed, that among all experiments with predictive models, those with LogP optimization showed the highest proportion of valid molecules and, at the same time, the predictive model for LogP estimation had the highest accuracy $R^2 = 0.91$ (see Methods). Probably it is harder for RL system to exploit high quality predictive model *P* and produce fictitious SMILES strings with predicted properties in the desired region.

**Table 1**. **Comparison of statistics for generated molecular datasets**

| Property | | Valid molecules, % | Mean SA score | Mean molar mass | Mean value of target property | Match with ZINC15 database, % | Match with ChEMBL database, % |
|---|---|---|---|---|---|---|---|
| Melting temperature | baseline | 95 | 3.1 | 435.4 | 181 | 4.7 | 1.5 |
| | minimized | 31 | 3.1 | 279.6 | 137 | 4.6 | 1.6 |
| | maximized | 53 | 3.4 | 413.2 | 200 | 2.4 | 0.9 |
| $pIC_{50}$ for JAK2 | baseline | 95 | 3.1 | 435.4 | 5.70 | 4.7 | 1.5 |
| | minimized | 60 | 3.85 | 481.8 | 4.89 | 2.5 | 1.0 |
| | maximized | 45 | 3.7 | 275.4 | 7.85 | 4.5 | 1.8 |
| log P | baseline | 95 | 3.1 | 435.4 | 3.63 | 4.7 | 1.5 |
| | range opt. | 70 | 3.2 | 369.7 | 2.58 | 5.8 | 1.8 |
| Number of benzene rings | baseline | 95 | 3.1 | 435.4 | 0.59 | 4.7 | 1.5 |
| | maximized | 83 | 3.15 | 496.0 | 2.41 | 5.5 | 1.6 |
| Number of substituents | baseline | 95 | 3.1 | 435.4 | 3.8 | 4.7 | 1.5 |
| | maximized | 80 | 3.5 | 471.7 | 7.93 | 3.1 | 0.7 |

**Model analysis.** Model interpretation is a highly significant component in any ML study. In this section we demonstrate how Stack-RNN learns and memorizes useful information from the SMILES string as it is being processed. More specifically, we have manually analyzed neuron gate activations of the neural network as it processes the input data.

Figure 6 lists several examples of cells in neural networks with interpretable gate activations. In this figure, each line corresponds to activations of a specific neuron at different SMILES processing time steps by the pre-trained baseline generative model. Each letter is colored according to the value of *tanh* activation in a cool-warm colormap from dark blue to dark red, i.e., from -1 to 1. We found that our RNN has several interpretable cells. These cells can be divided into two groups – chemically sensible groups, which activate in the presence of specific chemical groups or moieties, and syntactic groups, which keep tracks of numbers, bracket opening and closure, and even of SMILES string termination when the new molecule is generated. For instance, we saw cells reflecting the presence of a carbonyl group, aromatic groups or NH moieties in heterocycles. We also observed that in two of these three examples there were counter-cells that deactivate in the presence of the aforementioned chemical groups. Neural network-based models are notoriously uninterpretable (*54*) and the majority of cells were indeed in that category. On the other hand, the possibility of even partial interpretation offered by this approach could be highly valuable for a medicinal chemist.

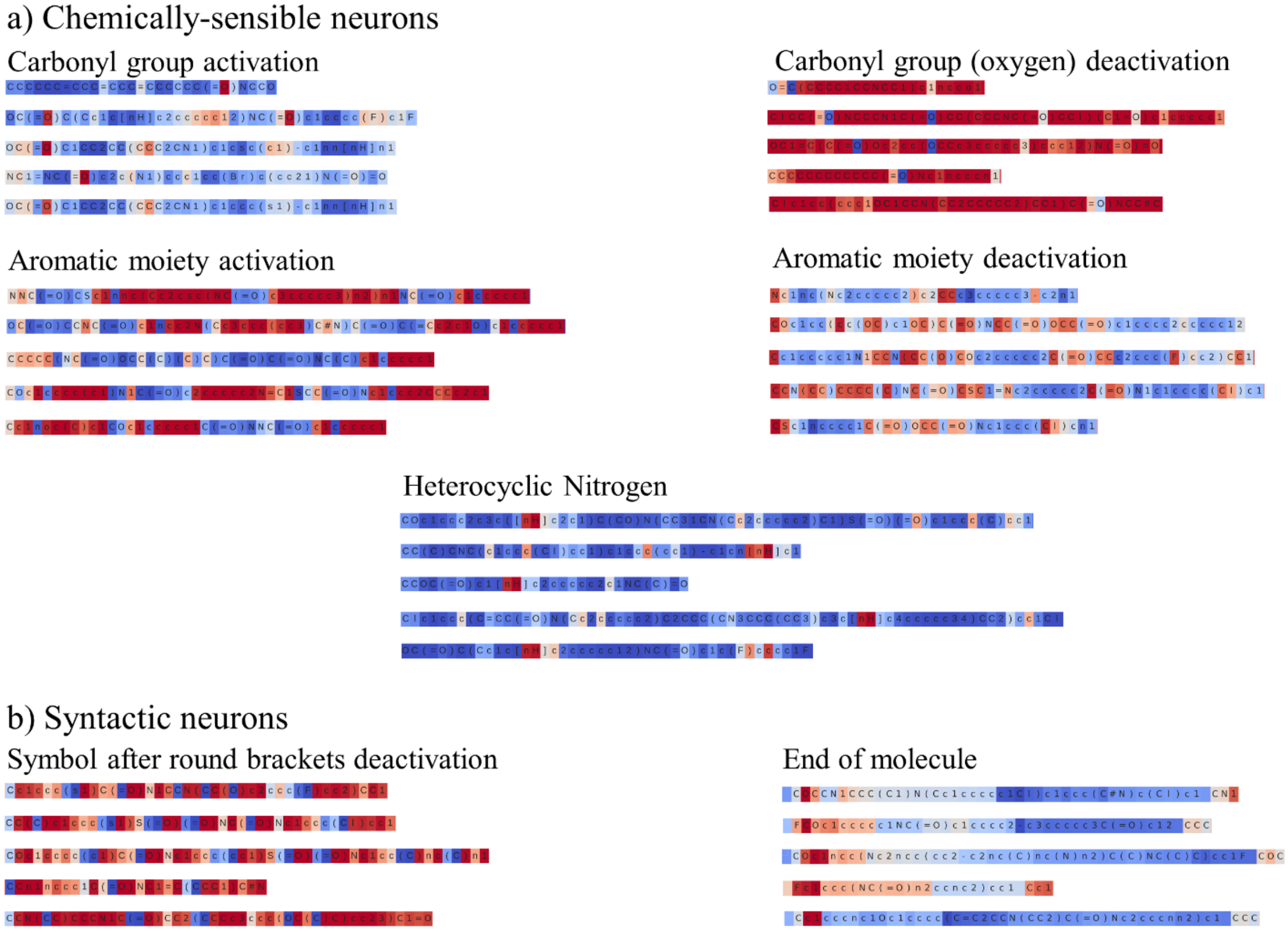

**Figure 6. Examples of Stack-RNN cells with interpretable gate activations.** Color coding corresponds to GRU cells with hyperbolic tangent $tanh$ activation function, where dark blue corresponds to the activation function value of -1 is and red described the value of the activation function of 1; the numbers in the range between -1 and 1 are colored using cool-warm color map.

**Visualization of new chemical libraries.** In order to understand how the generative models populate chemical space with new molecules, we used t-Distributed Stochastic Neighbor Embedding (t-SNE) for dimensionality reduction (*55*). We selected datasets for three endpoints used in our case studies ($T_m$, LogP, JAK2) that were produced with corresponding optimized generative models *G*. For every molecule, we calculated a latent vector of representation from the feed-forward layer with ReLU activation function in the predictive model *P* for the respective property and constructed 2D projection using t-SNE. These projections are illustrated in Figure 7. Every point corresponds to a molecule and is colored according to its property value.

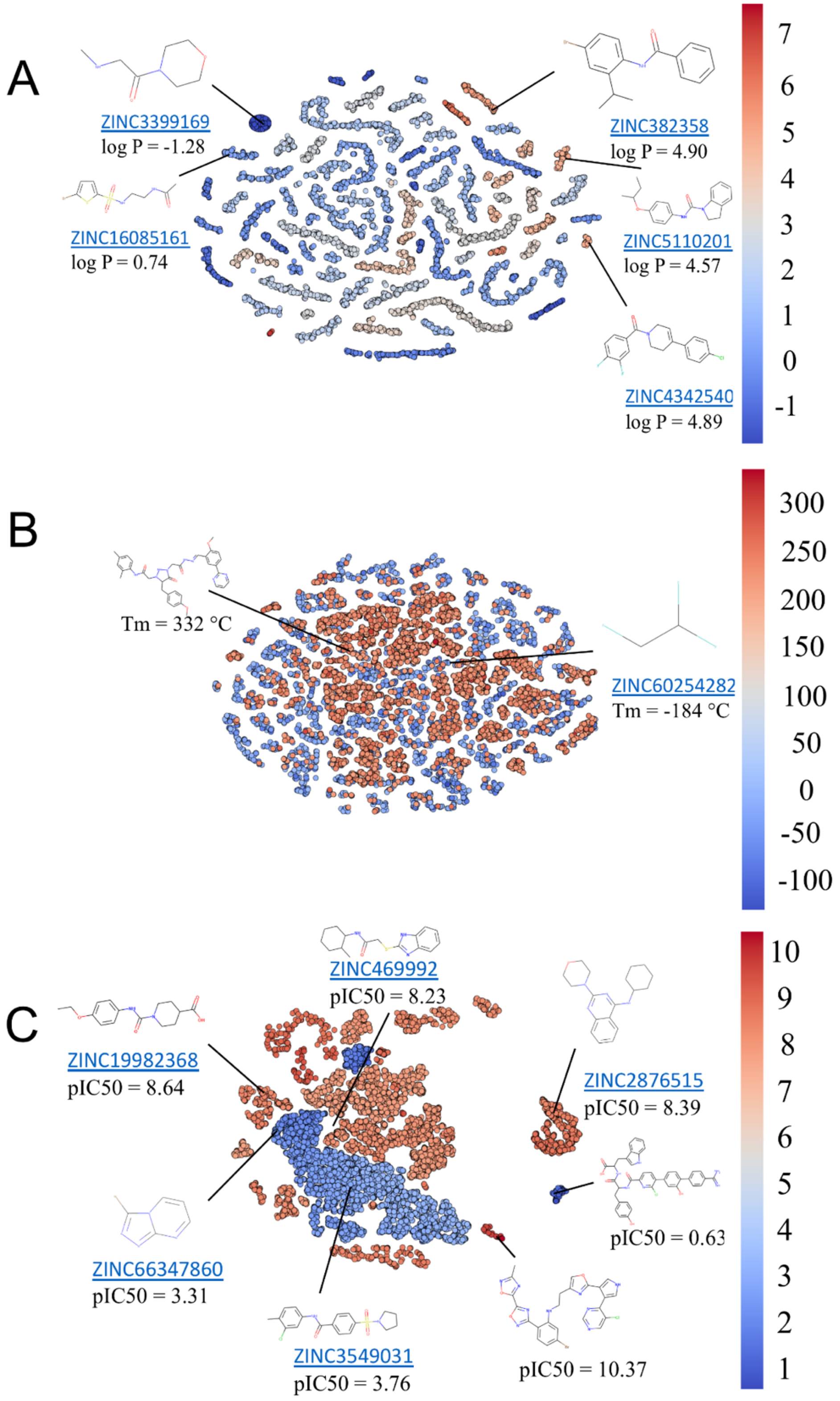

**Figure 7. Clustering of generated molecules by t-distributed stochastic neighbor embedding (t-SNE).** Molecules are colored based on the predicted properties by the predictive model *P*, with values shown by the color bar on the right. **(A, C)** Examples of the generated molecules are randomly picked from matches with ZINC database, property values predicted by the predictive model *P*. **(A)** Partition coefficient, logP **(B)** Melting

temperature, $T_m$ °C; examples show generated molecules with lowest and highest predicted melting temperatures; **(C)** JAK2 inhibition, predicted $pIC_{50}$.

For libraries generated to achieve certain partition coefficient distribution (Figure 7A), we can observe well-defined clustering of molecules with similar LogP values. In contrast, for melting temperature (Figure 7B) there are no such clusters. High and low $T_m$ molecules are intermixed together. This observation can be explained by the fact, that melting temperature depends not only on the chemical structure of the molecule, but also on intermolecular forces as well as packing in the crystal lattice. Therefore, plotting molecules in this neural net representation could not achieve good separation of high vs. low $T_m$. In the case of the JAK2 model, we could observe two large non-overlapping areas roughly corresponding to inactive ($pIC_{50}<6$) and active($pIC_{50}\geq6$) compounds. Inside these areas, molecules are typically clustered around multiple privileged scaffolds. Specifically for JAK2 we see abundance of compounds with 1,3,5-triazine, 1,2,4-triazine, 5-Methyl-1H-1,2,4-triazole, 7H-pyrrolo[2,3-d]pyrimidine, 1H-pyrazolo[3,4-d]pyrimidine, thieno-triazolo-pyrimidine and other substructures. Overall, this approach offers a rapid way to visualize compound distribution in chemical space in terms of both chemical diversity and variability in the values of the specific prediction endpoint. Furthermore, joint embedding of both molecules in the training set and those generated *de novo* allows one to explore differences in the chemical space coverage by both sets and establish whether structurally novel compounds also have the desired predicted property of interest.

## Discussion

We have created and implemented a deep reinforcement learning approach termed ReLeaSE for *de novo* design of novel chemical compounds with desired properties. To achieve this outcome, two deep neural networks – generative and predictive were combined in a general workflow that also included the RL step (Figure 1). The training process consists of two stages. In the first stage, both models are trained separately using supervised learning, and in the second stage, models are trained jointly with a reinforcement learning method. Both neural networks employ end-to-end deep learning. The ReLeaSe method does not rely on pre-defined chemical descriptors; the models are trained on chemical structures represented by SMILES strings only. This distinction makes this approach clearly differentiated from traditional QSAR methods and simpler both to use and execute.

This method needs to be evaluated in the context of several previous and parallel developments elsewhere to highlight its unique innovative features. Our ReLeaSE method has benefited from the recent developments in the machine learning community as applied to natural language processing and machine translation. These new algorithms allow learning the mapping from an input sequence (e.g., a sentence in one language) to an output sequence (that same sentence in another language). The entire input sentence represents an input vector for the neural network. The advantage of this approach is that it requires no handcrafted feature engineering.

Considering the use of similar approaches in chemistry, several comparable developments elsewhere should be discussed. Reinforcement Learning approach for *de-novo* molecular design was introduced in reference (*37*) as well. However, no data was provided to show that the predicted properties of molecular compounds are optimized. Instead of demonstrating the shift in distribution of biological activity values against dopamine receptor type 2 (DRD2) before and after the optimization, that study showed an increase in the fraction of the generated molecules, that are similar to training and test sets. This increase does not mean automatically, that the generative model is capable of producing novel active compounds. On the contrary, this result may indicate a model's weaknesses in predicting novel valuable chemicals that are merely similar to the training set compounds; that is the model is fitted to the training set but may have a limited ability to generate novel chemicals that are substantially different from the training set compounds. Indeed, the generative model in references (*23*, *37*) is a "vanilla" recurrent neural network without augmented memory stack, which does not have the capacity to count and infer algorithmic patterns(*34*). Another weakness of the approach described in reference (*37*), from our point of view, is the usage of a predictive model built with numerical molecular descriptors, whereas we propose a model, which is essentially descriptor-free and naturally forms a coherent workflow together with the generative model. After this manuscript was submitted for publication, a study by Jacques et al was published that used simple RNN and off-policy RL to generate molecules.(*56*) However, in addition to low percentage (~30-35%) of valid molecules, in that study, authors did not *directly* optimize any physical or biological properties but rather a proxy function that includes a synthetic accessibility score, drug likeness and a ring penalty.

It is important to highlight the critical element of using QSAR models as part of our approach as opposed to traditional use of QSAR models for virtual screening of chemical libraries. The absolute majority of compounds generated de novo by the ReLeaSE method are novel structures as compared to the datasets used to train generative models, and any QSAR model could be used to evaluate their properties. However, one of our chief objectives was to develop a method that can tune not only structural diversity (cf. Case study 1), but most importantly, bias the property (physical or biological) toward the desired range of values (case studies 2 and 3). The principal element of the ReLeaSE method as compared to traditional QSAR models is that QSAR models are implemented within the ReLEeaSE such as to put "pressure" on the generative model. Thus, albeit any QSAR model could evaluate properties of new chemicals, those built into our method are used directly for reinforcement learning to bias *de novo* library design toward the desired property.

As a proof of principle, we tested our approach on three diverse types of endpoints: physical properties, biological activity and chemical substructure bias. The use of flexible reward function enables different library optimization strategies where one can minimize, maximize or impose a desired range to a property of interest in the generated compound libraries. As a by-product of these case studies, we have generated a dataset of over 1M of novel compounds. In this work we have focused on presenting the new methodology and its application for initial hit generation. However, ReLeaSE could also be used for lead optimization, where a particular privileged scaffold is fixed and only substituents are optimized. Our future studies will explore this direction.

Computational library design methods are often criticized for their inability to control synthetic accessibility of de novo generated molecules (*13*). Indeed, computationally generated compounds are often quite complex; for instance, they may include exotic substituents. In many cases, such compounds may require multi-step custom syntheses or even could be synthetically inaccessible with the current level of technology. In the pharmaceutical industry, the ease of synthesis of a prospective hit molecule is of primary concern as it strongly affects the cost of the manufacturing process required for the industrial-scale production. For all experiments in this paper, the synthetic accessibility of de novo generated focused libraries was estimated using the SAS score (*41*). Distributions of SAS values are shown in Supplementary Figure S3, and the medians of the SAS scored are listed in Table 1. This analysis shows clearly that property optimization does not significantly affect synthetic accessibility of the generated molecules. The biggest shift of 0.75 for the distribution median was observed in the proof-of-concept study targeting the design of JAK2 inhibitors with minimized activity. Less than 0.5% of molecules had a high SAS of >6, which is an approximate cutoff for systems that are difficult to synthesize (*41*).

Obviously, it is technically feasible to include SAS score as an additional reward function; however, in our opinion, there are two main reasons as to why this is not desirable, at least with the current form of SAS. First, predicted SAS score for newly generated molecules is practically independent of property optimization. Its distribution follows that from commercially available compounds. Second, "synthetic accessibility" is not a well-defined concept (*57*). In the process chemistry it depends on multiple factors that determine the ease of synthesis of a particular molecule such as the availability of reagents, the number and difficulty of synthetic steps, the stability of intermediate products, the ease of their separation, reaction yields, etc (*58*). In contrast, the most commonly used SAS method (also used in this work) is based on molecular complexity as defined by the number of substructures and molecular fragments (*41*). Therefore, the optimizing SAS score with RL used in our approach will result in substantially reduced novelty of generated molecules and a bias toward substructures with low SAS scores used to train the model.

In summary, we have devised and executed a new strategy termed ReLeaSE for designing libraries of compounds with the desired properties that employs both deep learning and reinforcement learning approaches. In choosing the abbreviation for the name of the method, we were mindful of one of the key meanings of the word "release", i.e., to "allow or enable to escape from confinement; set free". We have conducted computational experiments that demonstrated the efficiency of the proposed ReLeaSE strategy in a single-task regime where each of the endpoints of interest is optimized independently. However, this system can be extended to afford multi-objective optimization of several target properties concurrently, which is the need of drug discovery where the drug molecule should be optimized with respect to potency, selectivity, solubility, and ADMET properties. Our future studies will address this challenge.

## Materials and Methods

**Experimental Data.** The melting point dataset was extracted from the literature (*50*). The PHYSPROP database (https://www.srcinc.com) used to extract the octanol/water partition coefficient, LogP for diverse set of molecules. Experimental $IC_{50}$ and $K_i$ data tested against JAK2 (CHEMBL ID 2971) was extracted from ChEMBL (*36*), PubChem (*59*) and Eidogen-Sertanty KKB (http://www.eidogen.com). Compounds that had inconclusive $IC_{50}$ values were considered unreliable and were not included in the modeling.

**Data curation.** Compiled datasets of compounds were carefully curated following the protocols proposed by Fourches et al. (*60*) Briefly, explicit hydrogens were added, and specifics chemotypes such as aromatic and nitro groups were normalized using ChemAxon Standardizer. Polymers, inorganic salts, organometallic compounds, mixtures, and duplicates were removed. Modeling-ready curated dataset contained 14,176 compounds for LogP, 15,549 compounds for JAK2 and 47,425 for melting temperature. All molecules were stored as normalized and canonicalized SMILES strings according to procedures developed elsewhere (*61*).

**Property prediction models.** We have built Quantitative Structure-Property Relationship (QSPR) models for three different properties – melting temperature, LogP and $pIC_{50}$ for JAK2. Curated datasets for all three endpoints were divided into training and training sets in five-fold cross-validation (5CV) fashion. In developing these QSPR models, we followed standard protocols and best practices for QSPR model validation (*42*). Specifically, it has been shown that multiple random splitting of datasets into training and test sets affords models of the highest stability and predictive power. Distinctly, models built herein did not use any calculated chemical descriptors; rather, SMILES representations were used. Each model consisted of an embedding layer transforming the sequence of discrete tokens (i.e., SMILES symbols) into a vector of 100 continuous numbers, LSTM layer with 100 units and *tanh* nonlinearity, one dense layer with 100 units and rectify nonlinearity function and one dense layer with one unit and identity activation function. All three models were trained with learning rate decay technique until convergence. The resulting 5CV external accuracies of the models are shown in Figure S4.

**Training for the generative model.** In the first stage, we pre-trained a generative model on a ChEMBL21 (*36*) dataset of approximately 1.5M drug-like compounds, so that the model was capable of producing chemically feasible molecules (note that this step does not include any property optimization). This network had 1500 units in a recurrent GRU (*32*) layer and 512 units in a stack augmentation layer. The model was trained on a GPU for 10000 epochs. The learning curve is illustrated in Figure S5.

The generative model has two modes of processing sequences – training and generating. At each time step, in the training mode, the generative network takes a current prefix of the training object and predicts the probability distribution of the next character. Then, the next character is sampled from this predicted probability distribution and is compared to the ground truth. Afterwards, based on this comparison the cross-entropy loss function is calculated and parameters of the model are updated. At each time step, in generating mode, the generative network takes a prefix of already generated sequences and then, like in the training mode, predicts the probability distribution of the next character and samples it from this predicted distribution. In the generative model, we do not update the model parameters.

At the second stage, we combine both generative and predictive model into one reinforcement learning system. In this system, the generative model plays the role of an agent, whose action space is represented by the SMILES notation alphabet and state space is represented by all possible strings in this alphabet. The predictive model plays the role of a critic estimating the agent's behavior by assigning a numerical reward to every generated molecule (i.e., SMILES string). The reward is a function of the numerical property calculated by the predictive model. At this stage, the generative model is trained to maximize the expected reward. The entire pipeline is illustrated in Figure 1.

We trained a stack-augmented RNN as a generative model. As mentioned above, for training we used the ChEMBL database of drug-like compounds. ChEMBL includes approximately 1.5 million of SMILES strings; however, we only selected molecules with the lengths of SMILES string of fewer than 100 characters. The length of 100 was chosen because more than 97% of SMILES in training dataset had 100 characters or less (see Figure S6).

**Stack-augmented recurrent neural network** (*30*)**.** This section describes generative model *G* in more details. We assume that the data is sequential, which means that it comes in the form of discrete tokens, i.e., characters. The goal is to build a model that is able to predict the next token conditioning on all previous tokens. The regular recurrent neural network has an input layer and a hidden layer. At time step $t$ neural network takes the embedding vector of token number *t* from the sequence as an input and models the probability distribution of the next token given all previous tokens, so that the next token can be sampled from this distribution. Information of all previously observed tokens is aggregated in the hidden layer. This can be written down as follows:

$$h_t = \sigma(W_i x_t + W_h h_{t-1}), \quad (6)$$

where $h_t$ is a vector of hidden states, $h_{t-1}$ – vector of hidden states from the previous time step, $x_t$ – input vector at time step $t$, $W_i$ – parameters of the input layers, $W_h$ -- parameter of the hidden layer and $\sigma$ – activation function.

The stack memory is used to keep the information and deliver it to the hidden layer at the next time step. A stack is a type of persistent memory, which can be only accessed through its topmost element. There are three operations supported by the stack: `POP` operation, which deletes an element from the top of the stack, `PUSH` operation, which puts a new element to the top of the stack; and `NO-OP` operation, which performs no action. The top element of the stack has value $s_t[0]$ and is stored at position 0:

$$s_t[0] = a_t[PUSH]\sigma(Dh_t) + a_t[POP]s_{t-1}[1] + a_t[NO-OP]s_{t-1}[0]. \quad (7)$$

where $D$ is $1 \times m$ matrix and $a_t = \left[a_t[PUSH], a_t[POP], a_t[NO-OP]\right]$ is a vector of stack control variables, which define the next operation to be performed. If $a_t[POP]$ is equal to 1, then the value below is used to replace the top element of the stack. If $a_t[PUSH]$ is equal to 1, then a new value will be added to the top and all the rest values will be moved down. If $a_t[NO - OP]$ equals 1 then stack keeps the same value on top.

Similar rule is applied to the elements of the stack at a depth *i>0*:

$$s_t[i] = a_t[PUSH]s_{t-1}[i-1] + a_t[POP]s_{t-1}[i+1] + a_t[NO - OP]s_{t-1}[i]. \tag{8}$$

Now the hidden layer $h_t$ is updated as:

$$h_t = \sigma(Ux_t + Rh_{t-1} + Ds_{t-1}^k), \tag{9}$$

where *D* is a matrix of size $m \times k$ and $s_{t-1}^k$ are the first *k* elements for the top of the stack at time step *t*-1.

## Acknowledgments

**Funding:**
We thank Stephen Cappuzzi for technical discussions and reading the manuscript prior the publication. A. T. and O. I. acknowledge support from the Eshelman Institute for Innovation award as well as the Department of Defense grant DOD-ONR (N00014-16-1-2311). M. P. acknowledges financial support from the Skolkovo Institute of Science and Technology. O.I. acknowledges Extreme Science and Engineering Discovery Environment (XSEDE) award DMR-110088, which is supported by National Science Foundation grant number ACI-1053575. We gratefully acknowledge the support and hardware donation from NVIDIA Corporation and express our special gratitude to Mr. Mark Berger.

**Author contributions:** A.T and O.I. devised and led this project and edited the final version of this manuscript. M.P. and O.I. developed and implemented the ReLeaSE method. M.P. wrote the manuscript with input from all authors.

**Competing interests:** The authors declare that they have no competing interests.

**Data and materials availability:** All data needed to evaluate the conclusions in the paper are present in the paper and/or the Supplementary Materials. Code, data and pre-trained models are available from GitHub (*https://github.com/isayev/ReLeaSE* ) as well as ZIP archive in Supplementary Materials. We alos reference PyTorch implementation of our methods and provide iPython notebook examples explaining how to run the experiments described in our paper. Additional data related to this paper may be requested from the authors.

## Supplementary Materials

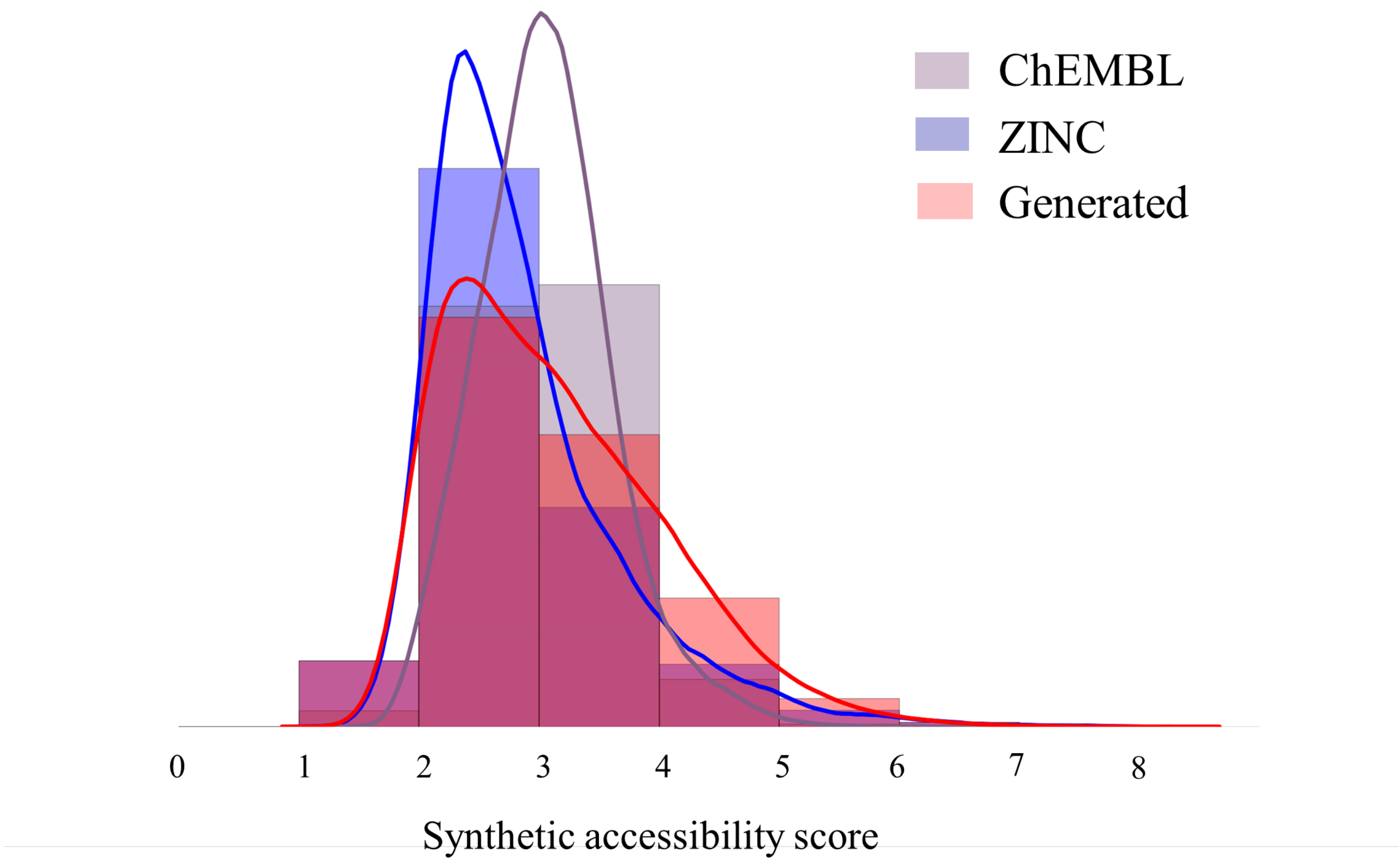

**Figure S1.** Distribution of Synthetic Accessibility Score (SAS) for the full ChEMBL21 database (~1.5M molecules), random subsample of 1M molecules from ZINC15 and generated dataset of 1M molecules with baseline generator model G.

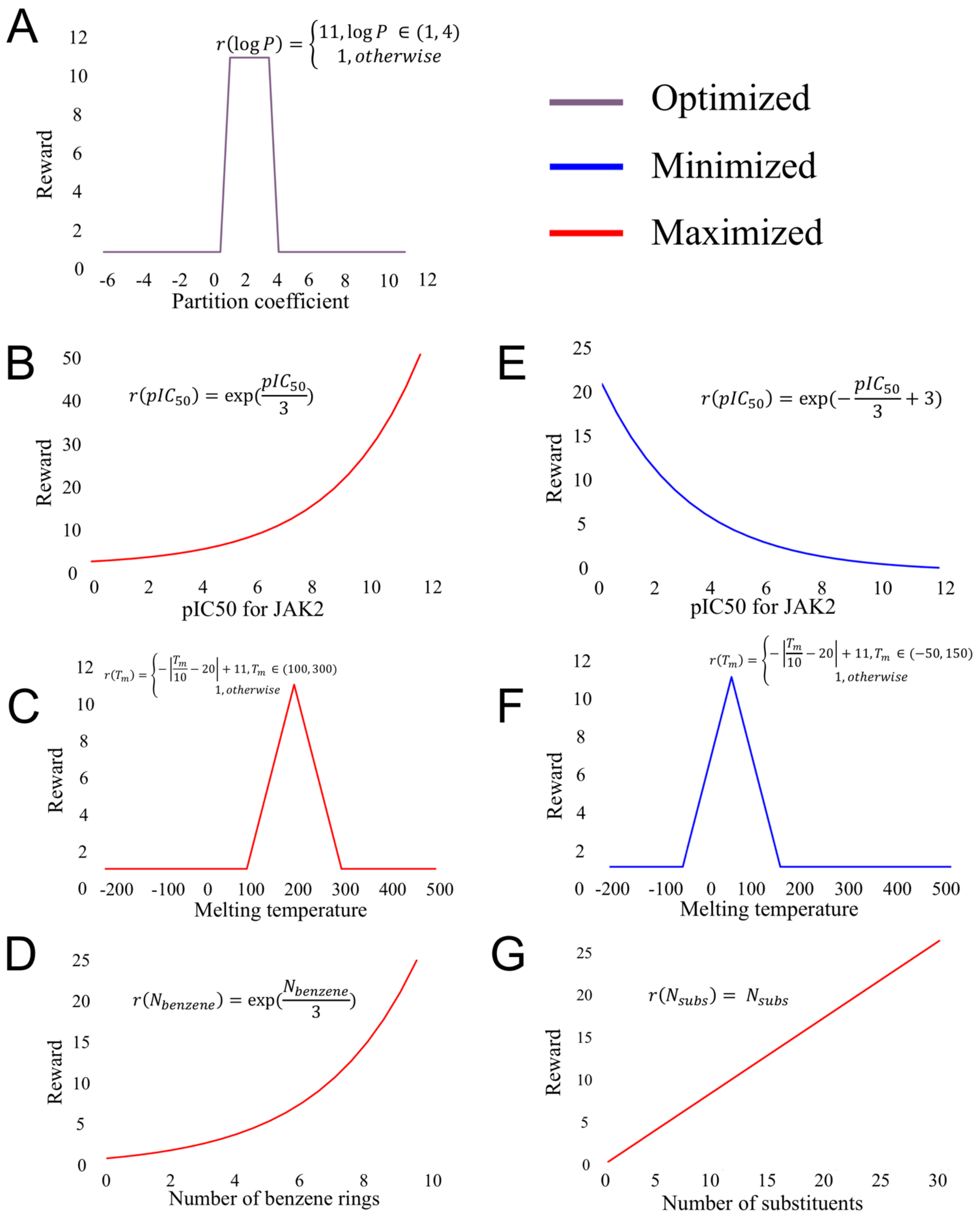

**Figure S2.** Reward functions. **(A)** logP optimization **(B)** $pIC_{50}$ for JAK2 maximization **(C)** Melting temperature maximization **(D)** Benzene rings maximization **(E)** $pIC_{50}$ for JAK2 minimization **(F)** Melting temperature minimization **(G)** Substituent maximization.

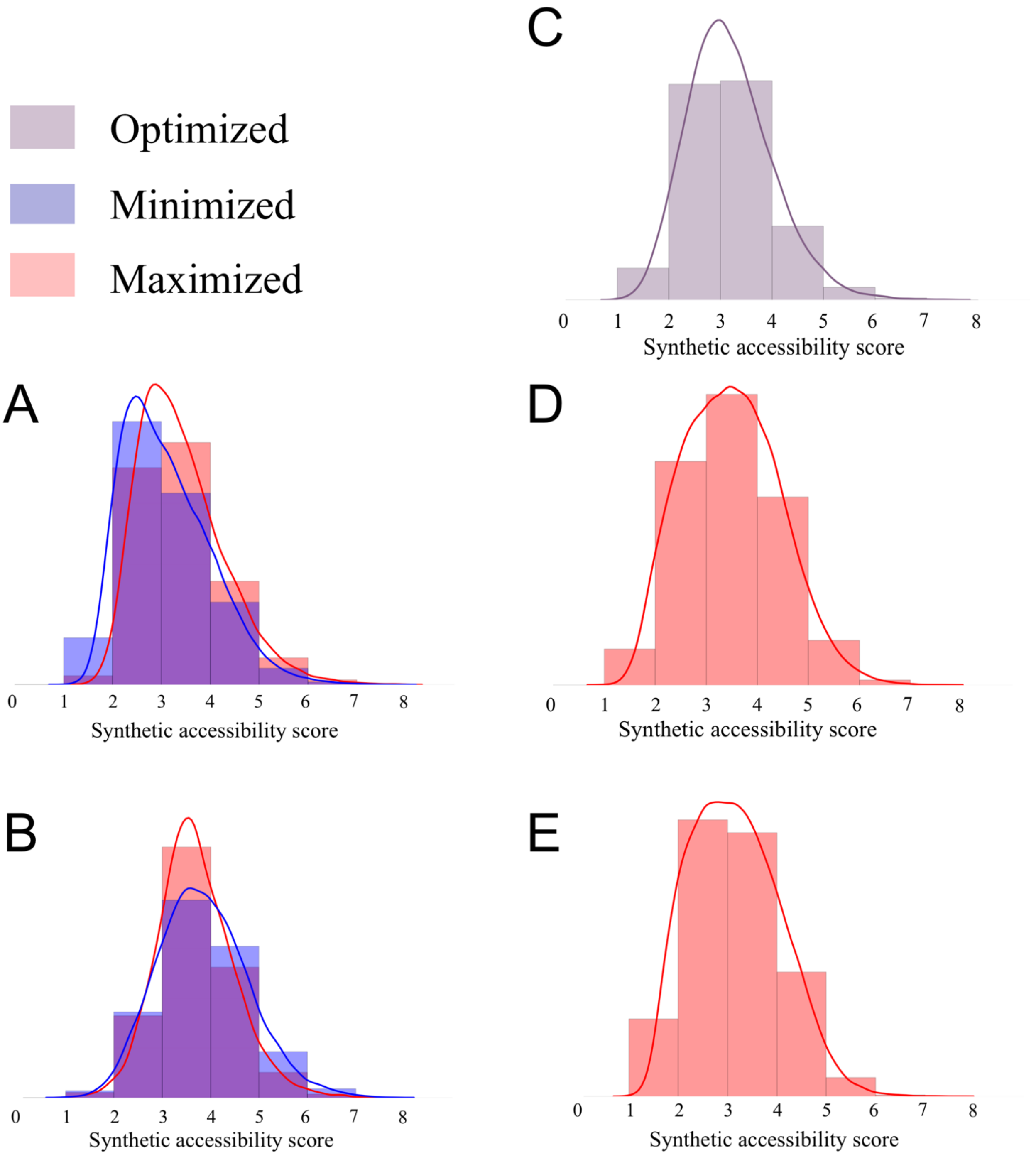

**Figure S3.** Distributions of Synthetic Accessibility Score for all RL experiments. **(A)** Melting temperature **(B)** JAK2 inhibition **(C)** Partition coefficient **(D)** Number of benzene rings **(E)** Number of substituents.

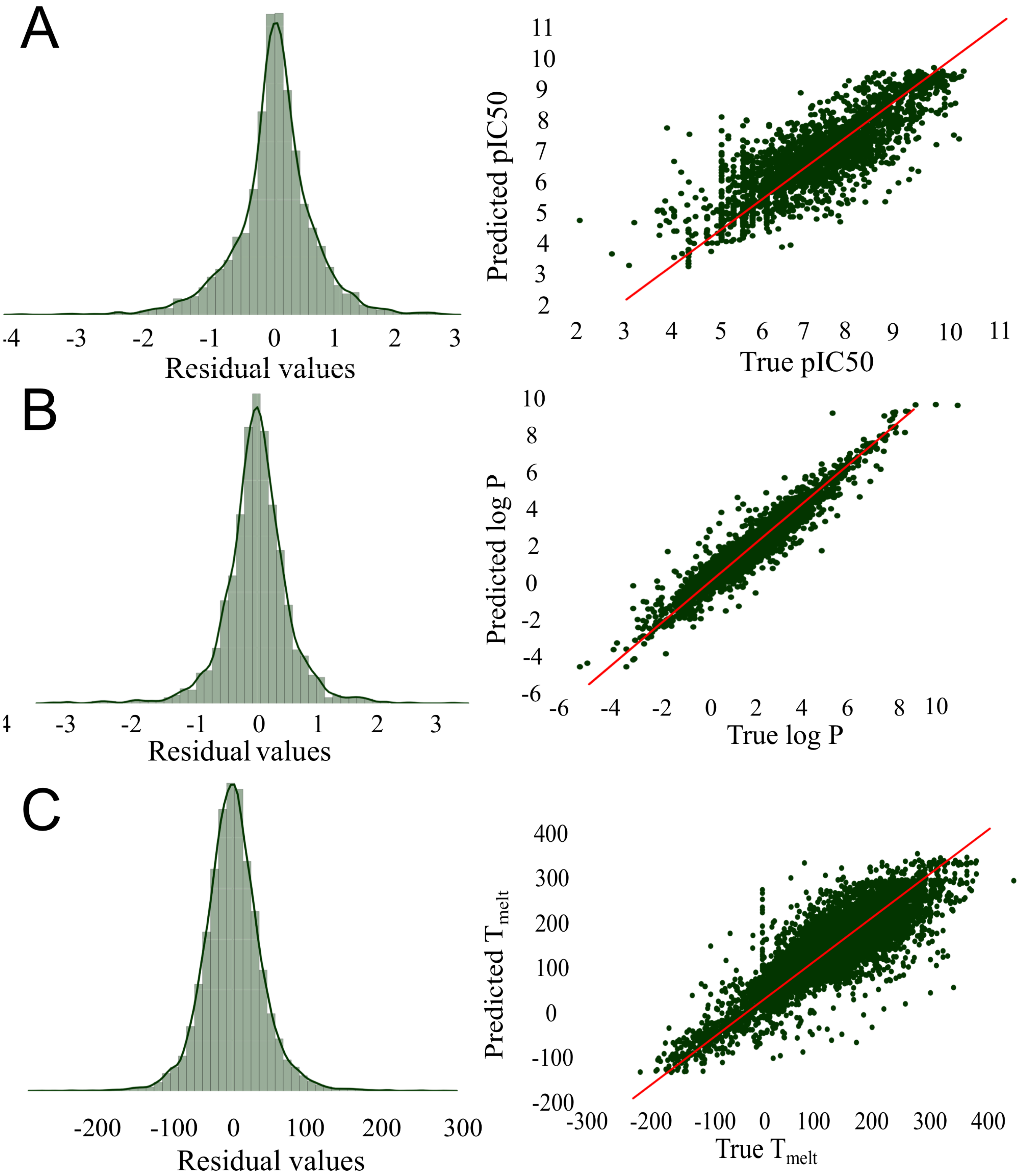

**Figure S4. Distribution of residuals and predicted vs. observed plots for predictive models.** Results are obtained with external Five-fold Cross-validation. All the values are calculated on hold out test datasets **(A)** Melting temperature **(B)** logP **(C)** $pIC_{50}$ for JAK2.

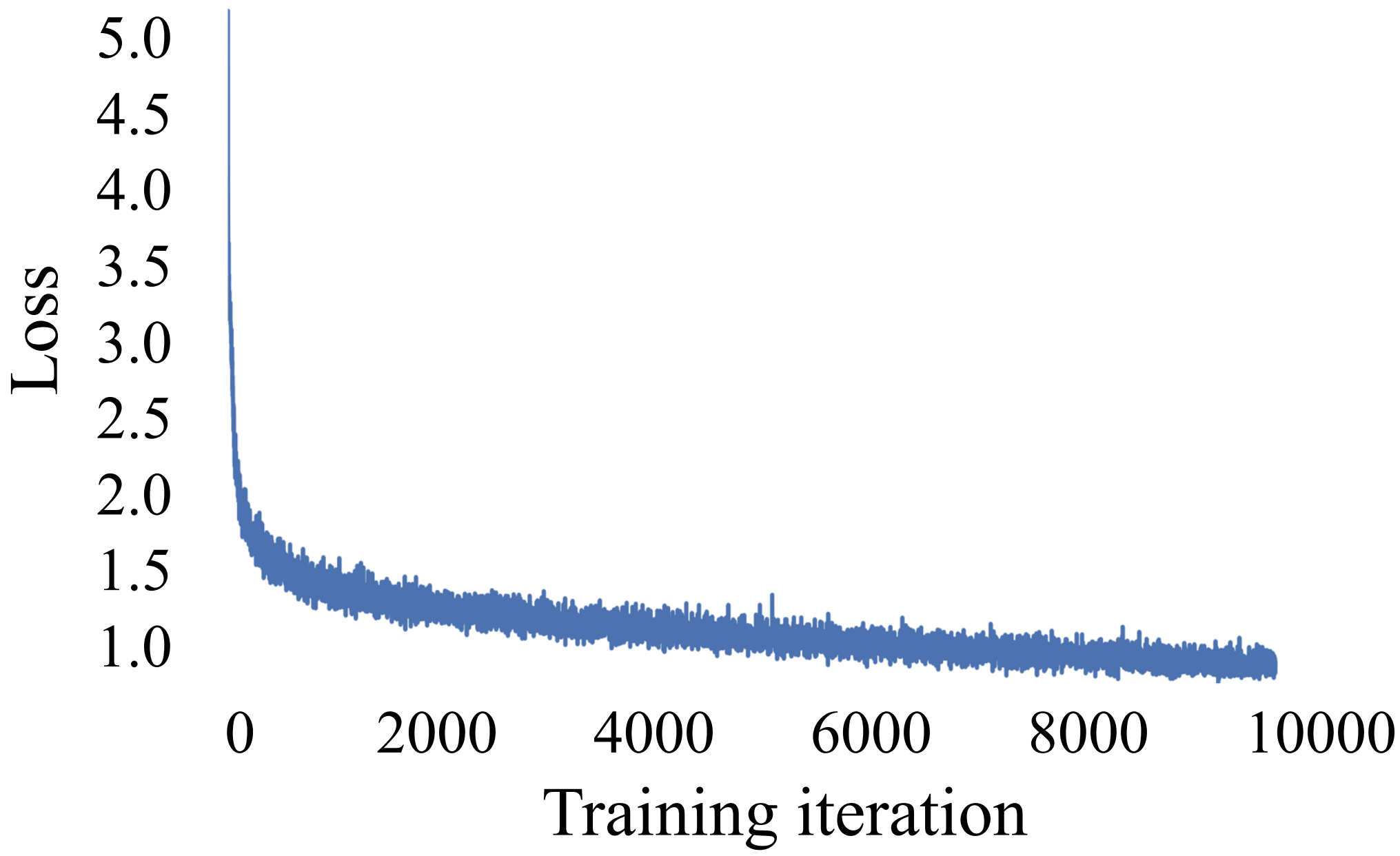

**Figure S5. Learning curve for generative model.**

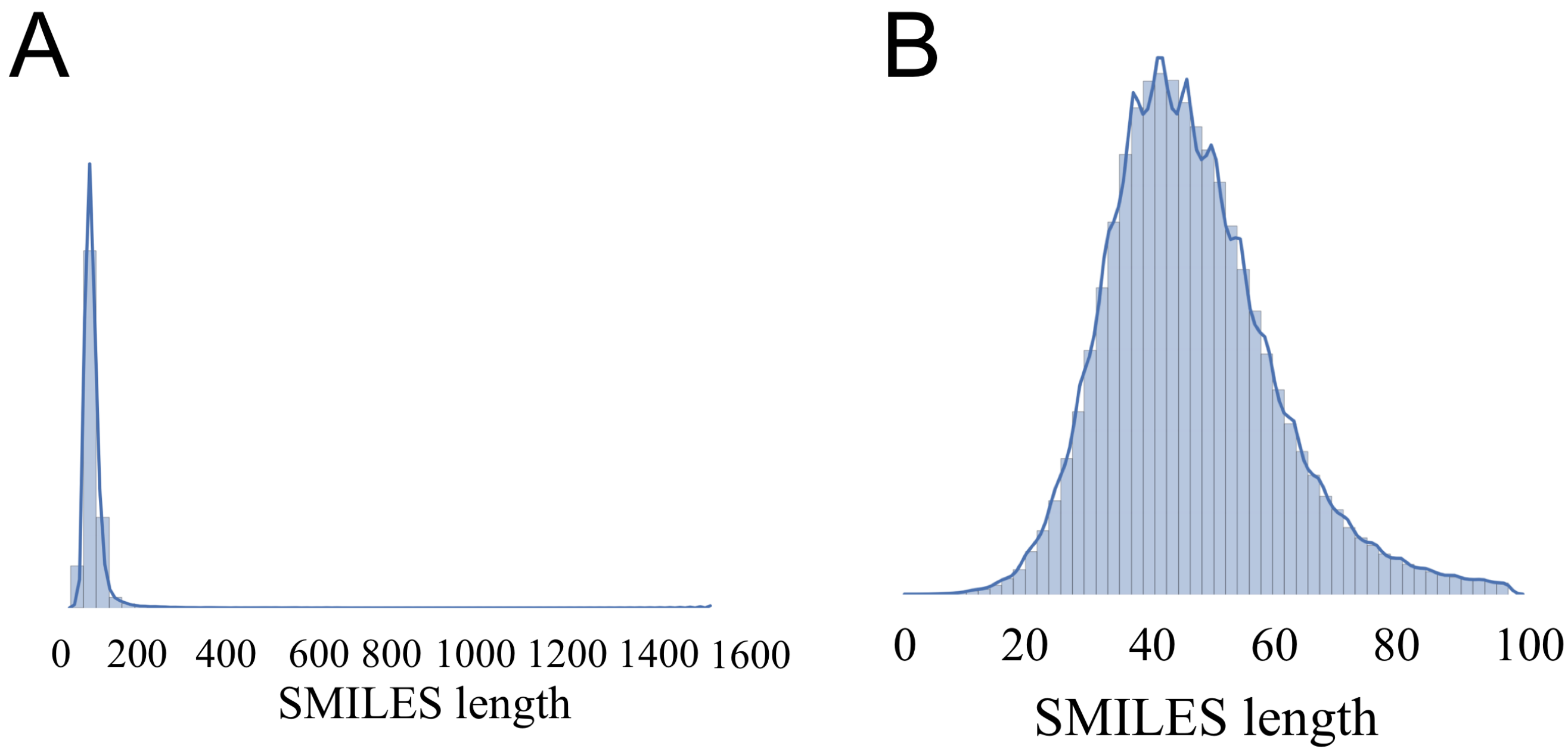

**Figure S6. Distributions of SMILES's strings lengths. (A)** Initial **(B)** Truncated.